\documentclass[letterpaper, 10 pt, conference]{ieeeconf} 
\IEEEoverridecommandlockouts                              
\usepackage{amsmath} 
\usepackage{amssymb}  
\usepackage{cite}
\usepackage{commath}
\usepackage{xcolor}
\definecolor{oiOrange}{HTML}{E69F00}
\definecolor{oiSkyBlue}{HTML}{56B4E9}
\definecolor{oiGreen}{HTML}{009E73}
\definecolor{oiYellow}{HTML}{F0E442}
\definecolor{oiBlue}{HTML}{0072B2}
\definecolor{oiVermillion}{HTML}{D55E00}
\definecolor{oiPurple}{HTML}{CC79A7}
\usepackage{array}
\usepackage[hidelinks]{hyperref}
\usepackage{tikz}
\usetikzlibrary{arrows.meta,decorations.pathmorphing, decorations.markings, calc, positioning, bending}
\usepackage{pgfplots}
\usepackage{comment}
\usepackage{url}
\usepackage{subcaption}
\newlength{\sublabelwidth}
\newcommand{\subfigwidth}{\dimexpr\columnwidth-\sublabelwidth\relax}
\newcommand{\sidesublabel}[2][0pt]{%
    \raisebox{#1}{%
        \begin{minipage}[c]{\sublabelwidth}
            \raggedright
            \rotatebox[origin=c]{90}{\parbox{2em}{\subcaption{}\label{#2}}}%
        \end{minipage}%
    }%
}

\DeclareMathAlphabet{\mathds}{U}{dsrom}{m}{n}
\newcommand{\ind}{\mathds{1}}

\newtheorem{definition}{Definition}

\DeclareMathOperator{\image}{im}

\DeclareMathOperator*{\argmin}{argmin}

\DeclareMathOperator{\subjto}{\text{subj. to}}
\DeclareMathOperator{\diag}{diag}

\title{\LARGE \bf
Orientation control of soft robots via adiabatic spectral submanifolds
}

\author{Aron Karakai, Roshan S. Kaundinya, Mike Y. Michelis, Robert K. Katzschmann, George Haller
\thanks{A. Karakai is with the Automatic Control Laboratory, ETH Zurich, Switzerland, {\tt\footnotesize akarakai@ethz.ch.} R. S. Kaundinya and G. Haller are with the Institute for Mechanical Systems, ETH Zurich, Switzerland, {\tt\footnotesize \{sroshan, georgehaller\}@ethz.ch.} M. Y. Michelis and R. K. Katzschmann are with the Soft Robotics Laboratory and the ETH AI Center, ETH Zurich, Switzerland, {\tt\footnotesize \{michelism, rkk\}@ethz.ch.}}%
}

\begin{document}

\maketitle

\begin{abstract}
Soft robots are commonly sought for safety-critical interactions in delicate environments, where accurate position and orientation control is imperative. Model predictive control (MPC) offers a solution, but it requires a model of the robot's infinite-dimensional nonlinear dynamics that is at once accurate and computationally cheap.  
Recent theory on adiabatic spectral submanifolds (aSSMs) and their applications to soft robots provide data-driven model-reduction methods to construct such models.
Here, we extend these methods to identify aSSMs from enlarged observable datasets and upgrade the currently available aSSM-MPC schemes. Evaluated on a high-fidelity finite-element simulation of a pressure-actuated soft arm, our controller reduces position and orientation tracking error by more than 60\% compared to existing data-driven baselines.
\end{abstract}

\section{Introduction}

As robotics evolves toward interacting with humans and nature, soft robots made of compliant, continuously deformable materials are gaining importance for navigating delicate and challenging environments. The design of soft robots often draws inspiration from biological systems, such as fish \cite{Katzschmann_DelPreto_MacCurdy_Rus_2018}, insects \cite{Ji_2019}, or the elephant trunk \cite{Toshimitsu_Wong_Buchner_Katzschmann_2021}. Naturally, accurate control of both position and orientation is crucial for the usefulness and safety of these flexible designs.

At the same time, such robots are typically built without explicit knowledge of the governing equations of their dynamics, which poses significant modeling challenges. That is because they have, in principle, infinitely many degrees of freedom, are underactuated, and exhibit intrinsically nonlinear behavior \cite{Della_Santina_Duriez_Rus_2023}. Thus, for most soft robots, only finite-element models with very high degrees of freedom are available at best, and the computational cost of those is prohibitively high for closed-loop control.

Due to these challenges, viable approaches to soft robot control are either based on directly mapping data to control inputs or on constructing a low-dimensional model that approximates the dynamics sufficiently well. Methods that belong to the former category are generally either rigorous techniques but limited to linear dynamics (see \cite{Coulson_Lygeros_Dorfler_2019} and \cite{van_Waarde_Eising_Camlibel_Trentelman_2023}), or black-box solutions, such as reinforcement learning (RL) \cite{Levine_Koltun_2013}, which learn a mapping from observations to control inputs. This often requires data collection on a scale that necessitates the use of an accurate simulator instead of the physical system, meaning that a high-fidelity model is needed in the first place (see \cite{pmlr-v229-jitosho23a} or \cite{Naughton_et_al_2021}).

Identifying a low-order model has two main paths. Firstly, approaches such as the piecewise constant curvature (PCC) or piecewise constant strain (PCS) approximations rely on simplifying assumptions about the physics to reduce the degrees of freedom; see \cite{Della_Santina_Duriez_Rus_2023} for an overview. However, these methods still require extensive knowledge of potentially unavailable information, such as the robot's stiffness, damping, and material properties.

\begin{figure}
    \centering
    \resizebox{0.625\linewidth}{!}{\begin{tikzpicture}[scale=1.0, line cap=round, line join=round, font=\huge]
    \coordinate (O) at (0,0);
    \draw[-{Stealth[length=3mm]}, very thick] (O) -- (5.0,0) node[right] {$x|_{E_\textrm{global}^\perp}$};
    \draw[-{Stealth[length=3mm]}, very thick] (O) -- (0,5.0) node[above] {$x|_{E_\textrm{global}}$};
    \draw[-{Stealth[length=3mm]}, very thick] (O) -- (-3.0,-3.0) node[below left] {$u$};

    \begin{scope}[yshift=-2cm]      

      \path[fill=oiGreen, fill opacity=0.12]
        (-2.25,-2.2)
          .. controls (-0.5, 0.1) and (1, 1.5) .. (2.8,2.75)
          .. controls (2.35,4.2) and (2.65,4.9) .. (2.8,6)
          .. controls (0.5, 4.5) and (-0.3, 4.3) .. (-2.25,2.3)
          .. controls (-2.5,0.6) and (-2,-0.5) .. (-2.25,-2.2)
          -- cycle;

      \draw[oiGreen, very thick, decoration={snake, amplitude=0.15mm, segment length=3mm, post length=0mm, pre length=0mm}, decorate]
        (-2.25,-2.2) .. controls (-2,-0.5) and (-2.5,0.6) .. (-2.25,2.3);

      \draw[oiGreen, very thick, decorate,
        decoration={snake, amplitude=0.6mm, segment length=6mm, post length=0mm, pre length=0mm}]
        (-2.25,2.3) .. controls (-0.3, 4.3) and (0.5, 4.5) .. (2.8,6);
      \draw[oiGreen, very thick, decoration={snake, amplitude=0.15mm, segment length=3mm, post length=0mm, pre length=0mm}, decorate]
        (2.8,2.75) .. controls (2.35,4.2) and (2.65,4.9) .. (2.8,6);
      \draw[oiGreen, very thick, decorate,
        decoration={snake, amplitude=0.6mm, segment length=6mm, post length=0mm, pre length=0mm}]
        (-2.25, -2.2) .. controls (-0.5, 0.1) and (1, 1.5) .. (2.8,2.75);

      \node[oiGreen] at (-2.9,1.7) {$\Tilde{\mathcal{A}}_\varepsilon$};
    \end{scope}
    \node[oiBlue] at (-3.2,-1.67) {$\mathcal{L}_\varepsilon$};

    \draw[oiBlue, very thick,
      decoration={markings, mark=at position 0.75 with {\arrowreversed{Stealth[length=3mm]}}}, decorate,
      decoration={snake, amplitude=0.2mm, segment length=3mm, post length=0mm, pre length=0mm}, decorate]
      (-3.2,-2.7)
        .. controls (-2.2,-1.2) and (-1.4,-0.7) .. (-0.6,-0.3)
        .. controls (0.03,-0.3) and (1.2,2.1) .. (3.2,2.3);

    \draw[oiVermillion, very thick, -{Stealth[length=3mm]},
      decoration={snake, amplitude=0.2mm, segment length=3mm, post length=3mm, pre length=0mm}, decorate]
      (5,3) .. controls (4.2,3) and (3.1,3) .. (2.2,2.8);
    \draw[oiVermillion, very thick,
      decoration={snake, amplitude=0.2mm, segment length=3mm, post length=1mm, pre length=0mm}, decorate]
      (-2,-1.25)
        .. controls (-1.65, -0.8) and (-1.1, -0.75) .. (-0.5, 0.6)
        .. controls (-0.1,2) and (1.0,2.5) .. (2.2,2.8);
    \node[oiVermillion] at (5.7,3.5) {$x(t; u^s(\varepsilon t) + u^d(t))$};
  \end{tikzpicture}}
    \vspace{-0.05cm}
    \caption{The (perturbed) aSSM $\Tilde{\mathcal{A}}_\varepsilon$ anchored to the slow manifold $\mathcal{L}_\varepsilon$ is a low-dimensional invariant manifold under inputs $u(t)=u^s(\varepsilon t) + u^d(t)$, where $u^s$ is slow and $u^d$ is uniformly bounded for all times.}
    \label{fig:assm}
    \vspace{-0.2cm}
\end{figure}
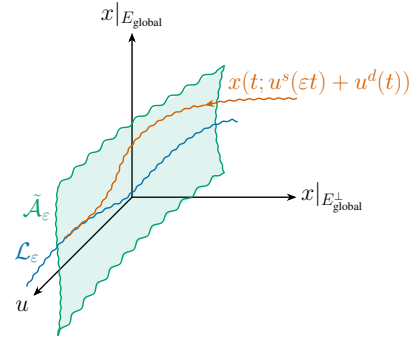

Secondly, a range of techniques aim to learn a low-dimensional model directly from data. The simplest methods for that involve fitting a linear dynamical system to recorded trajectory data. Dynamic mode decomposition with control (DMDc) \cite{Proctor_Brunton_Kutz_2016} and its extensions \cite{Proctor_Brunton_Kutz_2018}, applied to soft robots by \cite{Bruder_Fu_Gillespie_Remy_Vasudevan_2021} among others, implement this approach. However, fitting a linear model to an inherently nonlinear system lacks justification and generally yields low predictive power under inputs not seen during training.

For autonomous nonlinear systems, these limitations can be overcome using spectral submanifolds (SSMs) for data-driven model reduction \cite{Haller_Ponsioen_2016, Cenedese_Axas_Bauerlein_Avila_Haller_2022, Axas_Cenedese_Haller_2023}, capturing system dynamics on low-dimensional attracting invariant manifolds. This methodology was extended to aperiodic nonautonomous systems \cite{Haller_Kaundinya_2024} and recently deployed to soft robot control \cite{Kaundinya_Alora_Matt_Pabon_Pavone_Haller_2026} through adiabatic spectral submanifolds (aSSMs), leveraging a characteristic time-scale separation between fast internal dissipation in soft materials and comparatively slow tasks. While \cite{Kaundinya_Alora_Matt_Pabon_Pavone_Haller_2026} demonstrates the viability of aSSM-based position control of soft robots, its MPC formulation fixes both a static SSM approximation and a corresponding constant input sequence for each MPC planning horizon, optimizing only over fast but small deviations around the fixed inputs.

Here, we extend those methods further and develop an aSSM-MPC scheme that uses the full, time-varying aSSM geometry across the prediction horizon and optimizes over both slow and fast input components independently, thus improving predictive accuracy and control authority. We deploy this upgraded control method to the combined position and orientation control of the end effector of a soft continuum arm \cite{Toshimitsu_Wong_Buchner_Katzschmann_2021}. In high-fidelity finite-element simulations, our approach provides superior tracking performance compared to the available linear data-driven methods of \cite{Proctor_Brunton_Kutz_2016} and \cite{ Haggerty_Banks_Kamenar_Cao_Curtis_Mezic_Hawkes_2023} and significantly improves on the closed-loop accuracy for aSSM-reduced MPC controllers discussed in \cite{Kaundinya_Alora_Matt_Pabon_Pavone_Haller_2026}.

\textit{Contributions.} We deploy data-driven aSSM-reduction to model and control both the orientation and position of a pressure-driven soft arm. To that end, we
\begin{enumerate}
    \item extend the aSSM-MPC scheme of \cite{Kaundinya_Alora_Matt_Pabon_Pavone_Haller_2026} to use the full aSSM geometry and optimize over both slow and fast inputs,
    \item present a well-behaved set of orientation observables, and
    \item introduce a general-purpose Python implementation of data-driven aSSM-reduction\footnote{\url{https://github.com/karakaron/aSSMPy}}.
\end{enumerate}

\section{Problem setup}\label{sec:setup}
We consider soft robots with dynamics of the form
\begin{equation}\label{eq:system-ODE}
    \Dot{x}(t) = F(x(t), u(t)),
\end{equation}
where $x(t)\in\mathbb{R}^n$ is the state, $u(t)\in\mathcal{U}\subset\mathbb{R}^{n_u}$ is the control input from a compact set $\mathcal{U}$, and $F:\mathbb{R}^n \times \mathbb{R}^{n_u} \to \mathbb{R}^n$ is of class $C^r$ in its arguments for some $r\geq 1$.

In a realistic setting, $F$ is not known and the state $x$ cannot be measured directly. Instead, we have access to observables $y=\mu(x) \in \mathbb{R}^p$ and aim to control the workspace coordinates $z=c(y)\in\mathbb{R}^w$ of the robot, for some functions $\mu:\mathbb{R}^n\to \mathbb{R}^p$ and $c:\mathbb{R}^p\to \mathbb{R}^w$, where $w\leq p<<n$.

We consider control tasks in which a time-dependent target is to be tracked by the workspace coordinates. Such targets are denoted by $\Gamma(t) \in \mathbb{R}^w$ for $t\in[0, T_\Gamma]$, where $T_\Gamma>0$ is the duration of the task. A notable feature of soft robots is that they are highly dissipative, which results in fast internal decay rates. By contrast, the targets in typical applications have comparatively low velocities (see \cite{Fang_et_al_2021} or \cite{Ji_2019}). In line with that, we assume that $\Gamma$ has a slow time-dependence compared to the internal decay rates of the system \eqref{eq:system-ODE}.

In a data-driven setting, the time-scale separation can be quantified by a heuristic slowness metric, defined in \cite{Kaundinya_Alora_Matt_Pabon_Pavone_Haller_2026} as
\begin{equation}
    r_s = \Big(\tfrac{1}{T_\Gamma}\textstyle\int_0^{T_\Gamma}\norm{\Dot{\Gamma}(t)}dt\Big) \Big/ \Big(\tfrac{1}{T_d}\overline{\textstyle\int_0^{T_d} \norm{\Dot{z}_{\Bar{u}}(t)} dt}\Big),
\end{equation}
where $\Dot{z}_{\Bar{u}}$ stands for the time-derivative of the workspace coordinates $z$ under a static input $\Bar{u}\in\mathcal{U}$ and the overbar denotes the average taken over several such inputs. The trajectory length $T_d$ is chosen so that the typical decay properties of the system are captured. The aSSM-reduced models described in Section~\ref{sec:assm-reduction} are mathematically justified for $r_s << 1$. However, the experimental results of \cite{Kaundinya_Alora_Matt_Pabon_Pavone_Haller_2026} suggest that this can be relaxed to $r_s \lesssim 1$ in practice. Our experiments, see Section~\ref{sec:sopra-results}, lead to a similar observation.

\begin{figure}
    \centering
    \begin{subfigure}[B]{0.4\linewidth}
        \centering
        \includegraphics[width=3.3cm, angle=270]{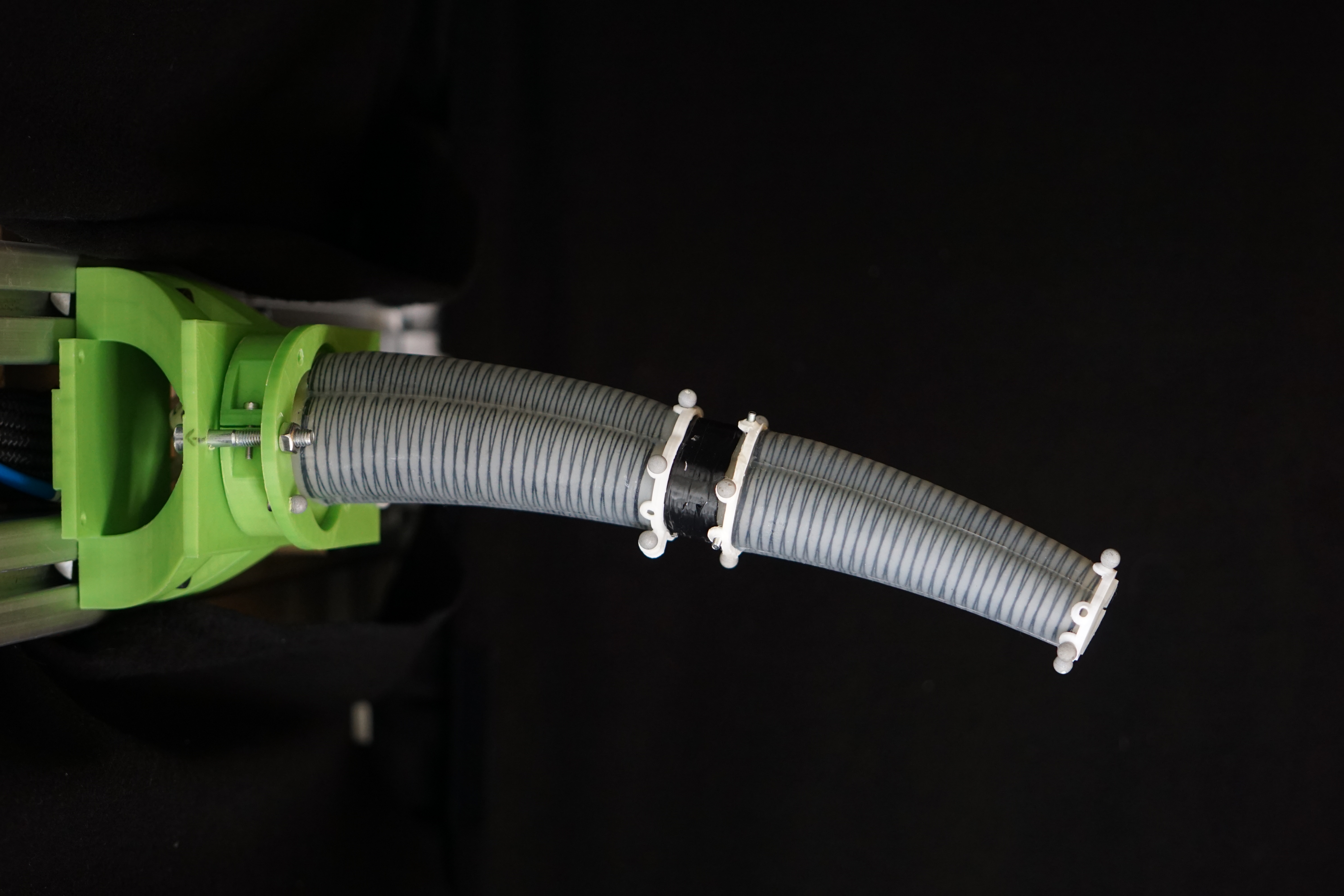}
    \end{subfigure}%
    \begin{subfigure}[B]{0.4\linewidth}
        \centering
        \includegraphics[height=3.3cm]{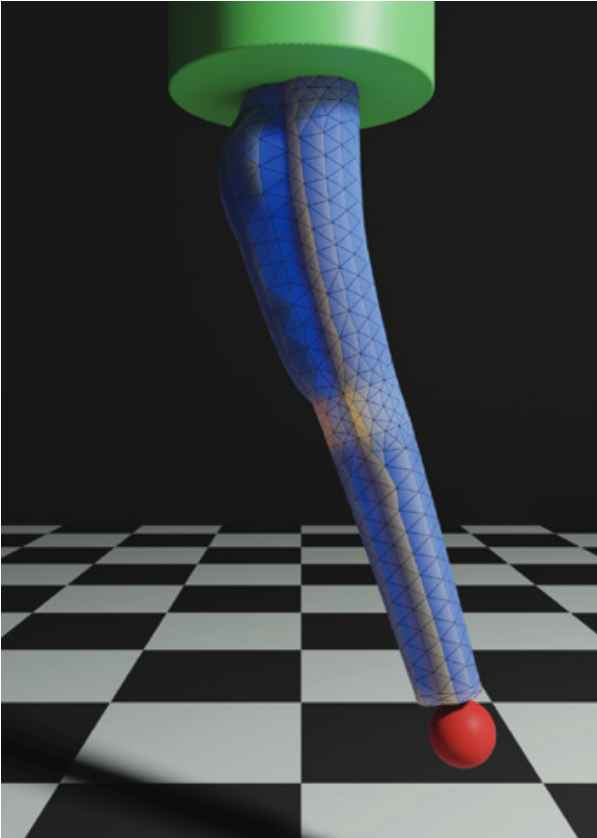}
    \end{subfigure}
    \caption{The SoPrA arm (left) and its model in the SORS simulator \cite{Mekkattu_Michelis_Katzschmann_2026} (right).}
    \label{fig:sopra}
    \vspace{-0.15cm}
\end{figure}

We aim to use the methods discussed in Sections~\ref{sec:assm-reduction} and \ref{sec:mpc} to model and control the soft arm SoPrA \cite{Toshimitsu_Wong_Buchner_Katzschmann_2021}, pictured in Figure~\ref{fig:sopra}. This is a 28 [cm] long soft arm, actuated with six pressure chambers. Its base is attached to a horizontal surface above the workspace, and it is hanging vertically when no input is applied.

To generate training data and test the MPC method, we use the SORS simulator \cite{Mekkattu_Michelis_Katzschmann_2026}. SORS is a high-fidelity soft robot simulator built on finite-element methods. We use its 3249-DOF SoPrA model with Young's modulus $100$ [kPa] and Poisson's ratio 0.4. 

\section{Data-driven model reduction via aSSMs}\label{sec:assm-reduction}
In this section, we review the model reduction of systems of the form \eqref{eq:system-ODE} through adiabatic spectral submanifolds, largely based on \cite{Kaundinya_Alora_Matt_Pabon_Pavone_Haller_2026}. We first introduce the theory assuming the dynamics \eqref{eq:system-ODE} are known, then discuss an approach for approximating the reduced-order model directly from data. 

\subsection{Adiabatic SSMs in control systems}\label{sec:assm-theory}
As discussed in Section~\ref{sec:setup}, our goal is to construct reduced-order models to be used in controllers for tracking slow targets. In an idealized scenario, a slow input would suffice for that. In reality, however, we need a fast feedback controller capable of making small corrections, in order to counteract model mismatch and external disturbances. Thus, as in \cite{Kaundinya_Alora_Matt_Pabon_Pavone_Haller_2026}, we consider inputs of the form
\begin{equation}\label{eq:adiabatic-input}
    u(t) = u^s(\varepsilon t) + u^d(t),
\end{equation}
where $0\leq \varepsilon << 1$ and $\norm{u^d(t)} \leq \delta$ with $0\leq \delta << 1$ uniformly in time.
That is, $u^s$ has a slow time-dependence, while $u^d$ is small but potentially fast.

System \eqref{eq:system-ODE} under an input of the form \eqref{eq:adiabatic-input} is a small perturbation from its $\delta=0$ limit, where $u(t)=u^s(\varepsilon t)$. Introducing $\alpha = \varepsilon t$ as a state, that limit becomes the autonomous slow-fast system
\begin{subequations}\label{eq:slow-input-extended}
    \begin{align}
        \Dot{x}(t) &= F(x(t), u(t)),\\
        \Dot{u}(t) &= \varepsilon D_\alpha u^s(\alpha),\\
        \Dot{\alpha} (t) &= \varepsilon,
    \end{align}
\end{subequations}
which is itself a perturbation from its $\varepsilon=0$ limit, where $u(t)=\Bar{u}\in\mathcal{U}$. We proceed by investigating that first, considering the family of autonomous systems
\begin{equation}\label{eq:static-u-ODE}
    \Dot{x}=F_{\Bar{u}}(x):=F(x, \Bar{u})
\end{equation}
for $\Bar{u}\in\mathcal{U}$, equivalent to \eqref{eq:system-ODE} with static inputs.

Throughout the paper, we assume that for all $\Bar{u}\in\mathcal{U}$, the dynamical system \eqref{eq:static-u-ODE} has a unique, asymptotically stable hyperbolic fixed point. This assumption is justified by the high energy dissipation in soft robots and allows us to define the manifold of fixed points as a graph over the admissible input set $\mathcal{U}$.

\begin{definition}
    The \textit{critical manifold} associated with the input set $\mathcal{U}$ is
    \begin{equation*}
        \mathcal{L}_0 = \{(x, \Bar{u}) \in \mathbb{R}^n\times \mathcal{U} \mid F(x, \Bar{u}) = 0\}.
    \end{equation*}
    Its parametrization as a graph over $\mathcal{U}$ is denoted by $S:\mathcal{U}\to \mathbb{R}^n$, so that $x=S(\Bar{u})$ for all $(x, \Bar{u})\in\mathcal{L}_0$.
\end{definition}

We proceed by constructing a reduced-order model for each autonomous system \eqref{eq:static-u-ODE}. To that end, we need to choose the dimension $d$ of these models such that the dynamics \eqref{eq:static-u-ODE} exhibit $d$ dominant directions in the state space. That is, we pick $d$ such that there is a spectral gap between the $d$ slowest eigenvalues $\lambda_1(\Bar{u}), \lambda_2(\Bar{u}), \dots, \lambda_d(\Bar{u})$ of the linearization $A(\Bar{u})=D_x F_{\Bar{u}}(x)\rvert_{x=S(\Bar{u})}$ and the rest, uniformly across static inputs $\Bar{u}\in \mathcal{U}$.

Defining $E(\Bar{u})$ as the spectral subspace associated with $\lambda_1(\Bar{u}), \lambda_2(\Bar{u}), \dots, \lambda_d(\Bar{u})$, this strict spectral splitting implies that $E(\Bar{u})$ is an invariant subspace of the linearized system $\Dot{x}=A(\Bar{u})x$ that contains its dominant long-term dynamics. The theory of spectral submanifolds \cite{Haller_Ponsioen_2016, Cabre_et_al_2003} extends this idea to the nonlinear dynamics \eqref{eq:static-u-ODE} by considering nonlinear continuations of slow spectral subspaces.

\begin{definition}
    Fix $\Bar{u}\in\mathcal{U}$ and consider the autonomous dynamical system \eqref{eq:static-u-ODE}. For $E(\Bar{u})$ as above, a \textit{spectral submanifold} (SSM) anchored to $S(\Bar{u})$ is a $d$-dimensional invariant manifold $\mathcal{W}(E(\Bar{u}))$, embedded in $\mathbb{R}^n$, which is tangent to $E(\Bar{u})$ at the fixed point $S(\Bar{u})$. An SSM of \eqref{eq:static-u-ODE} is called a \textit{static} SSM of the nonautonomous system \eqref{eq:system-ODE}.
\end{definition}

Under our assumptions and mild nonresonance conditions, \cite[Theorem~3]{Haller_Ponsioen_2016} guarantees the existence of a unique SSM $\mathcal{W}(E(\Bar{u}))$ of smoothness class $C^r$ at each steady state $S(\Bar{u})\in\mathcal{L}_0$. Moreover, these have a $C^r$-smooth dependence on $\Bar{u}$.
Taking the SSM $\mathcal{W}(E(\Bar{u}))$ anchored to each fixed point, we construct
\begin{equation}\label{eq:A_0}
    \mathcal{A}_0 = \bigcup_{\Bar{u}\in\mathcal{U}} \mathcal{W}(E(\Bar{u})).
\end{equation}

The manifold $\mathcal{A}_0$ is a $C^r$-smooth invariant manifold of the system \eqref{eq:slow-input-extended} in the $\varepsilon=0$ limit. Moreover, it is normally attracting, meaning that the normal attraction rates to $\mathcal{A}_0$ overpower the tangential attraction rates within $\mathcal{A}_0$.

Hence, the dynamics on $\mathcal{A}_0$ capture the dominant long-term behavior of the system \eqref{eq:system-ODE} with a static input. What remains is to generalize this construction to inputs of the form \eqref{eq:adiabatic-input}. Firstly, under slow variation of the input, that is, for $u(t)=u^s(\varepsilon t)$ with $\varepsilon$ sufficiently small, \cite[Theorem~5]{Haller_Kaundinya_2024} guarantees the existence of a one-dimensional attracting \textit{slow manifold} $\mathcal{L}_\varepsilon$ that is $C^r$-diffeomorphic and $\mathcal{O}(\varepsilon)$ $C^1$-close to the restriction of the critical manifold $\mathcal{L}_0$ to the input set traversed by $u^s$.

Secondly, as proven in \cite[Theorem~6]{Haller_Kaundinya_2024} and applied to systems with control inputs in \cite{Kaundinya_Alora_Matt_Pabon_Pavone_Haller_2026}, the restriction of $\mathcal{A}_0$ to the slow input trajectory diffeomorphically persists as an invariant manifold $\mathcal{A}_\varepsilon$. Moreover, $\mathcal{A}_\varepsilon$ is $\mathcal{O}(\varepsilon)$ $C^1$-close to $\mathcal{A}_0$ and contains $\mathcal{L}_\varepsilon$. We call $\mathcal{A}_\varepsilon$ an \textit{adiabatic spectral submanifold} (aSSM) anchored to $\mathcal{L}_\varepsilon$.

In its present form, the theory in the relevant literature does not guarantee that aSSMs are also normally attracting. However, in the sequel, we assume that to be the case. See point 9 of \cite[Remark~3.3]{Eldering_2013} for a reasoning in support of this assumption. Then, \cite[Theorem~3.1]{Eldering_2013} guarantees that $\mathcal{A}_\varepsilon$ perturbs into an invariant manifold $\Tilde{\mathcal{A}}_\varepsilon$ upon the addition of $u^d(t)$ to the input for sufficiently small $\delta > 0$, such that $\Tilde{\mathcal{A}}_\varepsilon$ is diffeomorphic and $\mathcal{O}(\delta)$ $C^1$-close to $\mathcal{A}_\varepsilon$.

\subsection{Learning aSSMs from data}\label{sec:data-driven}
An aSSM-based reduced-order model consists of three ingredients. Namely, a projection from the full state to reduced coordinates, an expression for the dynamics in the reduced coordinates, and a lifting map from reduced coordinates to the aSSM in the state space.

Since the full state $x\in\mathbb{R}^n$ cannot be measured in most applications, we aim to construct an aSSM-reduced model of the evolution of the observables $y\in\mathbb{R}^p$.
To ensure that the SSMs $\mathcal{W}(E(\Bar{u}))$ embed into the observable space, we use delay-embedding to increase the dimension of the latter. For ease of notation, we redefine $y$ and $p$ to stand for the delay-embedded vector of observables and its dimension, respectively. Then, a generalization \cite{Deyle_Sugihara_2011} of the Takens Embedding Theorem \cite{Takens_1981} guarantees that the images of the SSMs under generic observables are embedded submanifolds of the observable space $\mathbb{R}^p$, whenever $p\geq 2d + 1$.

The starting point for data-driven identification of aSSMs is the construction of static SSMs $\mathcal{W}(E(\Bar{u}))$ anchored to steady states associated with a random set of static inputs. To that end, we sample $N_u$ static inputs from $\mathcal{U}$ and record observables of $N_t$ decaying trajectories subject to each of these static inputs. After removing initial transients, the recorded trajectories are stored in matrices
\begin{equation}
    Y_i = \begin{bmatrix}
        y_i(0) & y_i(\Delta t) \hspace{0.55em} \cdots \hspace{0.55em} y_i((M-1) \Delta t)
    \end{bmatrix} \in \mathbb{R}^{p\times M},
\end{equation}
$i=1, 2, \dots, N_u N_t$, where $\Delta t$ is the sampling time and $M$ is the number of recorded time steps. Steady states are estimated by $\Bar{y}_i = y_i((M-1)\Delta t)$. Let $\Bar{Y}_i = \ind_{M}^\top \otimes \Bar{y}_i$ and
\begin{equation}\label{eq:Y-nonautonomous}
    Y = \begin{bmatrix}
        Y_1 - \Bar{Y}_1 & Y_2 - \Bar{Y}_2 & \cdots & Y_{N_u N_t} - \Bar{Y}_{N_u N_t}
    \end{bmatrix},
\end{equation}
so that $Y\in\mathbb{R}^{p\times N_u N_t M}$ collects all trajectories shifted to decay to zero. We similarly construct a data matrix of the associated fixed points as
\begin{equation}
    \Bar{Y} = \begin{bmatrix}
        \Bar{Y}_1 & \Bar{Y}_2 & \cdots & \Bar{Y}_{N_u N_t} 
    \end{bmatrix} \in \mathbb{R}^{p \times N_u N_t M}.
\end{equation}

The approximation of the critical manifold $\mathcal{L}_0$ from such data is a simple regression task. Namely, we fit a function $S:\mathcal{U} \to \mathbb{R}^p$ that maps static inputs to the associated fixed points in the observable space.
To be able to compute a reference input for an aSSM-MPC scheme, we also fit a map from steady states in the workspace to the corresponding static inputs. Note, however, that the latter are generally not unique, so $S^\dagger$ is only a right-inverse of $c\circ S$.

To learn the spectral subspaces $E(\Bar{u})$, we rely on a global approximation. In particular, we compute a subspace $E_\textrm{global}=\image V$ such that the columns of $V\in\mathbb{R}^{p\times d}$ are the $d$ dominant left singular vectors of $Y$. Then, $E_{\textrm{global}}$ consists of the $d$ longest-surviving directions of variation in the recorded trajectories. This is inspired by the procedure developed in \cite{Axas_Cenedese_Haller_2023} for autonomous systems.
We then approximate each $E(\Bar{u})$ by shifting $E_\textrm{global}$ with $S(\Bar{u})$ so that it is anchored to the appropriate steady state on $\mathcal{L}_0$. We take the coordinates on $E(\Bar{u})$ with respect to the basis collected in $V$ as reduced coordinates. That is, the projection of an observable $y$ to reduced coordinates is defined as
\begin{equation}
    \eta = V^\top (y - S(\Bar{u})) \in \mathbb{R}^d,
\end{equation}
using an orthogonal projection.

We fit a polynomial parametrization of $\mathcal{W}(E(\Bar{u}))$ as a graph over the spectral subspace $E(\Bar{u})$ for each static input $\Bar{u}$ in the training data. By including the fixed points $\Bar{y}$ as regression features, this fit extends to unseen static inputs in $\mathcal{U}$, thereby approximating $\mathcal{A}_0$. Formally, we solve
\begin{equation}\label{eq:W-regression}
    \hat{W}^\star = \argmin_{\hat{W}\in\mathbb{R}^{p\times m_w}} \hspace{1em} \norm{Y - \hat{W} \begin{bmatrix}
        V^\top Y\\
        \Bar{Y}
    \end{bmatrix}_{0:n_w^y}^{1:n_w^\eta}}^2,
\end{equation}
where $([\eta^\top \hspace{0.7em} \Bar{y}^\top]^\top)_{0:n_w^y}^{1:n_w^\eta}$ denotes the vector of monomials of order 1 to $n_w^\eta$ in $\eta$ and 0 to $n_w^y$ in $\Bar{y}$, with dimension denoted by $m_w$, meant columnwise in \eqref{eq:W-regression}. The choice of polynomial orders $n_w^\eta$ and $n_w^y$ is application-specific; see \cite{Kaundinya_Alora_Matt_Pabon_Pavone_Haller_2026} for details.

Using the same notation, the reduced dynamics are learned by fitting a polynomial expression in $\eta$ and $\Bar{y}$ to approximate $\Dot{\eta}$ under an input $\Bar{u}$. That is obtained by solving
\begin{equation}\label{eq:R-regression}
    \hat{R}^\star = \argmin_{\hat{R}\in\mathbb{R}^{d\times m_r}} \hspace{1em} \norm{V^\top \Dot{Y} - \hat{R} \begin{bmatrix}
        V^\top Y\\
        \Bar{Y}
    \end{bmatrix}_{0:n_r^y}^{1:n_r^\eta}}^2,
\end{equation}
where $n_r^\eta$ and $n_r^y$ are again chosen polynomial regression orders and $\Dot{Y}$ is computed with finite-differencing.
With $\hat{W}^\star$ and $\hat{R}^\star$ computed, we define the functions $W:\mathbb{R}^d\times\mathcal{U} \to \mathbb{R}^p$ and $R:\mathbb{R}^d\times\mathcal{U} \to \mathbb{R}^d$ by
\begin{equation*}
    W(\eta, u) = \hat{W}^\star \hspace{-0.2em} \begin{bmatrix}
        \eta\\
        S(u)
    \end{bmatrix}_{0:n_w^y}^{1:n_w^\eta} \hspace{0.3em} \text{and} \hspace{0.5em} R(\eta, u) = \hat{R}^\star \hspace{-0.2em} \begin{bmatrix}
        \eta\\
        S(u)
    \end{bmatrix}_{0:n_r^y}^{1:n_r^\eta}.
\end{equation*}

Lastly, the effects of the control deviation $u^d(t)$ on the reduced dynamics need to be learned. We record the observables of an additional trajectory for each static input $\Bar{u}$ in the training data, with uniformly bounded random noise added to these inputs. With that, we learn the first-order expansion of the reduced dynamics in the input around $\Bar{u}$. To be precise, we seek an effective linear control matrix $B:\mathcal{U} \to \mathbb{R}^{d\times n_u}$ such that $\Dot{\eta} \approx R(\eta, \Bar{u}) + B(\Bar{u})u^d$ for $u(t) = \Bar{u} + u^d(t)$. Such $B$ can be found through a regression problem similar to \eqref{eq:R-regression}, first order in $u^d$ and order $n_b$ in $S(\Bar{u})$.

Then, for inputs of the form \eqref{eq:adiabatic-input}, a data-driven reduced-order model is obtained by replacing $\Bar{u}$ with $u^s(\varepsilon t)$ in the expressions above. In particular, we get the model
\begin{subequations}\label{eq:rom}
\begin{align}
        \eta(0) &= V^\top (y(0) - S(u^s(0))),\\
        \Dot{\eta}(t) &= R(\eta(t), u^s(\varepsilon t)) + B(u^s(\varepsilon t))u^d(t), \label{eq:rom-dynamics}\\
        y(t) &= W(\eta(t), u^s(\varepsilon t)) + S(u^s(\varepsilon t)),
\end{align}
\end{subequations}
expressing the projection to reduced coordinates, the reduced dynamics, and the lifting from reduced coordinates to the perturbed aSSM $\Tilde{\mathcal{A}}_\varepsilon$ embedded in the observable space, respectively. The validity of this model follows from the results discussed in Section~\ref{sec:assm-theory} for small $\varepsilon$ and $\delta$.

This data-driven model reduction procedure can be carried out using our new Python implementation aSSMPy.

\section{MPC with aSSM-reduced models}\label{sec:mpc}
In this section, we introduce an MPC method that uses aSSM-reduced models for prediction, enabling the data-driven control of soft robots. Similar but simpler approaches to soft robot control can be found in the literature. In particular, \cite{Alora_Cenedese_Haller_Pavone_2025} and \cite{Alora_Cenedese_Schmerling_Haller_Pavone_2023} use an SSM anchored to a fixed point of the unforced system to control a soft robot with MPC around that point. To track targets further away in the workspace, an aSSM-based strategy is proposed by \cite{Kaundinya_Alora_Matt_Pabon_Pavone_Haller_2026}. Here, we extend that method further.

Let $t_k$ denote discretized time with step size $\Delta t$ and $N$ the planning horizon. Then, at each time step, MPC computes the control input by minimizing the predicted deviation of the workspace coordinates $z$ from the target $\Gamma$ over the next $N$ steps. For a comprehensive overview, see \cite{Rawlings_Mayne_Diehl_2017}.

Our MPC scheme uses the aSSM-reduced model \eqref{eq:rom} for predictions. In particular, let $\varphi(t; t_0, \eta_0, u^s(\cdot), u^d(\cdot))$ denote the solution of \eqref{eq:rom-dynamics} with initial condition $\eta(t_0)=\eta_0$ and input $u(t)=u^s(\lfloor t/\Delta t\rfloor \Delta t) + u^d(\lfloor t/\Delta t\rfloor \Delta t)$. Then, predictions are computed as
\begin{subequations}\label{eq:dynamics-constraints}
    \begin{align}
        \eta(t_k)&=V^\top(y(t_k) - S(u^s(t_{k-1}))),\\
        \eta(t_j)&= \varphi(t_j; t_k, \eta(t_k), u^s(\cdot), u^d(\cdot)),\\
        \Tilde{y}(t_j) &= W(\eta(t_j), u^s(t_{j-1})) + S(u^s(t_{j-1})),\\
        \Tilde{z}(t_j) &= c(\Tilde{y}(t_j)),
    \end{align}
\end{subequations}
for $j\geq k$, where $\Tilde{y}$ and $\Tilde{z}$ denote predicted observables and workspace coordinates, respectively, as opposed to measured values $y$ and $z$ of these variables.

With that, our MPC formulation is defined through the following optimization problem at each time step $t_k$:
\begin{subequations}\label{eq:assm-mpc}
    \begin{align}
        & \min_{\substack{u^s(t_k), \dots, u^s(t_{k+N-1}) \\ u^d(t_k), \dots, u^d(t_{k+N-1})}} && \norm{\Tilde{z}(t_{k+N}) - \Gamma(t_{k+N})}_{Q_f}^2 \nonumber \\
        &  && \hspace{-9em} + \sum_{j=k}^{k+N-1} \bigg(\norm{\Tilde{z}(t_{j}) - \Gamma(t_{j})}_{Q}^2 + \norm{\begin{bmatrix} u^s(t_{j}) - S^\dagger(\Gamma(t_j)) \\ u^d(t_{j}) \end{bmatrix}}_R^2 \nonumber\\
        & && \hspace{-9em} + \norm{\begin{bmatrix} u^s(t_{j}) - u^s(t_{j-1}) \\ u^d(t_{j}) - u^d(t_{j-1}) \end{bmatrix}}_{R_{du}}^2 \bigg)
    \end{align}
    \begin{align}
        & \subjto && \hspace{0em} \text{dynamics \eqref{eq:dynamics-constraints}}, \label{eq:mpc-reduced-dyn}\\
        & && \hspace{0em} u(t_j) = u^s(t_j) + u^d(t_j), \label{eq:mpc-input}\\
        & && \hspace{0em} u(t_j) \in \mathcal{U}, \hspace{0.6em} u^s(t_j) \in \mathcal{U}^s, \hspace{0.6em} u^d(t_j)\in\mathcal{U}^d, \label{eq:mpc-input-constraints}\\
        & && \hspace{0em} u^s(t_j) - u^s(t_{j-1}) \in d\hspace{0.025em}\mathcal{U}^s, \label{eq:mpc-slew-rate-constraints1}\\
        & && \hspace{0 em} u^d(t_j) - u^d(t_{j-1}) \in d\hspace{0.025em}\mathcal{U}^d, \label{eq:mpc-slew-rate-constraints2}\\
        & && \hspace{0em} \Tilde{z}(t_j) \in \mathcal{Z}, \hspace{1em} \Tilde{z}(t_{k+N}) \in \mathcal{Z}_f, \label{eq:mpc-terminal-constraints}
    \end{align}
\end{subequations}
where $j$ runs from $k$ to $k+N$ in \eqref{eq:mpc-reduced-dyn} and to $k+N-1$ in \eqref{eq:mpc-input}-\eqref{eq:mpc-terminal-constraints}. The matrices $Q, Q_f, R_{du}\succeq 0$, and $R\succ 0$ are tunable weights, following standard MPC notation. The constraints enforce the dynamics for the predictions and that the inputs are chosen from the appropriate sets.
The remaining constraints \eqref{eq:mpc-slew-rate-constraints1}-\eqref{eq:mpc-terminal-constraints} can be used to enforce slowness of the inputs and safety in the workspace.

Although the aSSM-based MPC scheme developed by \cite{Kaundinya_Alora_Matt_Pabon_Pavone_Haller_2026} is the main inspiration behind our method, it takes a slightly different approach. Namely, at the start of each planning horizon, it fixes the slow input at $S^\dagger(z(t_k))$ for the entire horizon and uses the corresponding static SSM for predictions. That is, it sets $u^s$ so that it renders the current configuration a fixed point and only optimizes in $u^d$. 

By contrast, we treat both the slow input sequence $u^s(t_k), \dots, u^s(t_{k+N-1})$ and the input deviation sequence $u^d(t_k), \dots, u^d(t_{k+N-1})$ as independent optimization variables and use the full aSSM geometry for predictions. This allows the slow inputs to vary within a planning horizon, meaning that the inputs applied to the system are chosen with greater freedom, while the full aSSM geometry yields more accurate trajectory predictions to optimize over.

At the same time, we can design the parameters $R_{du}=\diag(R_{du}^s, R_{du}^d)$ of the slew rate cost to reward the slowness of $u^s$ and, similarly, tune $R=\diag(R^s, R^d)$ to penalize large $u^d$, independent of the corresponding cost of the other input variable. The same applies to input constraints \eqref{eq:mpc-input-constraints} and slew rate constraints \eqref{eq:mpc-slew-rate-constraints1}-\eqref{eq:mpc-slew-rate-constraints2}. Hence, the input assumptions of the aSSM-reduced model can be explicitly encouraged or enforced.
Lastly, while \cite{Kaundinya_Alora_Matt_Pabon_Pavone_Haller_2026} sets the reference input to zero, we plug in an approximate open-loop strategy for tracking $\Gamma$, computed via the critical manifold, thus reducing steady-state error.

Our aSSM-MPC scheme is implemented using GuSTO \cite{Bonalli_Cauligi_Bylard_Pavone_2019}, a sequential convex programming technique that solves \eqref{eq:assm-mpc} by iteratively approximating it with QPs.

\section{Control of a soft arm with aSSM-MPC}\label{sec:sopra-results}

Our goal is to control the orientation and position of the end effector of SoPrA, the soft arm introduced in Section~\ref{sec:setup}. This robot is actuated through six pressure chambers, three along the sides of
its bottom half, and a similar three in the top half. Inputs are the desired pressures achieved using a low-level PID controller at each chamber.

With this alignment, the pressure chambers cannot twist the arm. Hence, to control orientation, we propose adding a DC motor at the base of the robot, where it is mounted, to rotate the entire arm. To enforce the existence of a unique fixed point for each static input, as assumed in Section~\ref{sec:assm-theory}, we equip the DC motor with a PD controller that drives the base rotation to a target angle by applying appropriate currents. We model this low-level PD controller and the PID controllers of the pressure chambers as part of the system.

Here, we consider a rotation range of $-0.5$ to $0.5$ radians for the DC motor at the base. The pressures are limited to 75 [kPa] in the lower and 65 [kPa] in the upper segment of the arm, dictated by constraints of the simulator. We scale the corresponding inputs to the unit interval, which yields the input set $\mathcal{U}=[-0.5, 0.5]\times [0,1]^6$.

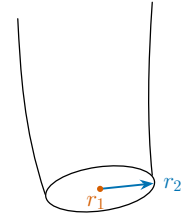
\begin{figure}
    \centering
    \resizebox{0.27\linewidth}{!}{\begin{tikzpicture}[scale=0.6, line cap=round, line join=round, font=\Large]
    \coordinate (C) at (0,0);        
    \coordinate (P) at (1.95,0.2);   

    \draw[thick, black, line cap=round]
      (-2,-0.2)
        .. controls (-2.8,2.2) and (-2.9,4.3) .. (-3.0,6.2);
    \draw[thick, black, line cap=round]
      (1.9,0.5)
        .. controls (1.7,2.4) and (1.75,4.6) .. (1.9,6.8);

    \draw[thick, black] (C) ellipse [x radius=2, y radius=0.8, rotate=8];

    \draw[oiBlue, very thick, -{Stealth[length=3mm]}] (C) -- (P);
    \node[oiBlue, right] at ($(P)+(0.15,-0.05)$) {$r_2$};

    \fill[oiVermillion] (C) circle (3.2pt);
    \node[oiVermillion, below] at ($(C)+(-0.15,-0.05)$) {$r_1$};
  \end{tikzpicture}}
    \caption{Observables of SoPrA.}
    \label{fig:observables}
    \vspace{-0.2cm}
\end{figure}

As observables, we take a position vector $r_1\in\mathbb{R}^3$ and an orientation vector $r_2\in\mathbb{R}^3$. Specifically, $r_1$ encodes the coordinates of the center of the end effector, relative to the center of the base, and $r_2$ points from the center of the end effector to a chosen point on its edge. See Figure~\ref{fig:observables}. This is motivated by the description of rigid body rotations with the Lie group $SO(3)$. In particular, for a rigid arm with a ball joint at the base, $r_1$, $r_2$, and $r_1\times r_2$ could be identified, up to scaling, with the columns of the rotation matrix in $SO(3)$ that uniquely describes the arm's motion. While this argument does not hold precisely for a soft arm, the expressivity of these observables regarding the end effector's dynamics carries over. These observables can also be recorded using a motion capture system \cite{Cangan_et_al_2022}. As opposed to using angles, this choice avoids the need for modular arithmetic in the data-driven modeling process and in MPC.

We record these observables following the procedure set out in Section~\ref{sec:data-driven}. In particular, we use the parameters $N_u=150$, $N_t=3$, $M=600$, and $\Delta t=10$ [ms]. The random inputs are generated with Latin hypercube sampling \cite{mckay1979} and are applied sequentially with steep ramp transitions, reshuffled after each pass over the random input set. The first 10 points, including the ramp transitions and initial transients, are removed from each trajectory.

For learning the critical manifold, we extend the dataset with 100 additional random inputs. We use $10\%$ Perlin noise \cite{Perlin1985} for the noisy inputs needed for fitting $B$. In total, this amounts to 70 minutes of training data.

Our experiments suggest two independent oscillations, namely the $x$ and $y$ coordinates of the position observable, and three overdamped modes, corresponding to the rotation of the arm and the vertical motion of the end effector, as the dominant dynamics. Hence, we use an SSM dimension of $d=7$. Indeed, the linearization of the resulting model at the unforced fixed point has the eigenvalues $\lambda_1 = -2.43$, $\lambda_2=-2.81$, $\lambda_3=-4.84$, $\lambda_{4,5}=-7.68\pm32.2i$, and $\lambda_{6,7}=-7.71\pm32.78i$. To satisfy the Takens Embedding Theorem, we use two delays, leading to $p=18$. 

We use the polynomial approximation orders $n_w^\eta=n_r^\eta=2$ and $n_w^y = n_r^y=n_b=1$, chosen to balance performance and computational cost.

\subsection{Control performance}
We test the aSSM-based MPC scheme developed in Section~\ref{sec:mpc}, with the model discussed above, on two test tasks. These tasks comprise dynamic targets for the position $r_1$ of the end effector and the first two coordinates of the orientation vector $r_2$, noting that the tilt of the end effector relative to the $x-y$ plane cannot be controlled. That is, the workspace coordinates are $z=(r_{1_x}, r_{1_y}, r_{1_z}, r_{2_x}, r_{2_y})\in\mathbb{R}^5$.

\begin{figure}
    \centering
    \sidesublabel[-0.0260\subfigwidth]{fig:task1-pos}%
    \begin{minipage}[c]{\dimexpr\columnwidth-\sublabelwidth\relax}
        \centering
        \includegraphics[width=\linewidth]{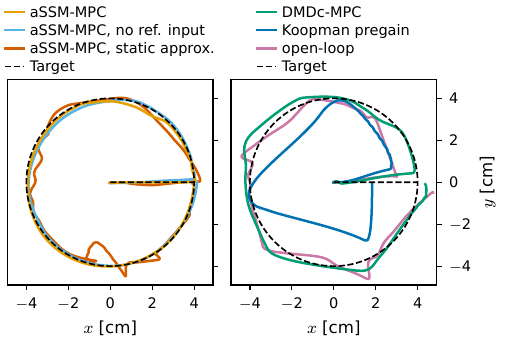}
    \end{minipage}

    \sidesublabel[0.0552\subfigwidth]{fig:task1-angle}%
    \begin{minipage}[c]{\dimexpr\columnwidth-\sublabelwidth\relax}
        \centering
        \includegraphics[width=\linewidth]{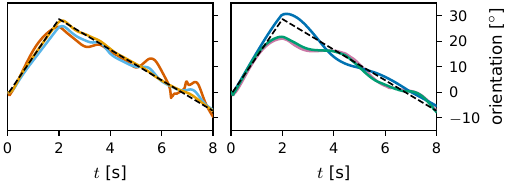}
    \end{minipage}
    \vspace{-0.12cm}
    \caption{Position (a) and orientation (b) of the end effector in task 1. Orientation is plotted as the angle $\arctan(z_5/z_4)$.}
    \label{fig:task1}
    \vspace{-0.15cm}
\end{figure}

\begin{figure}
    \centering
    \sidesublabel[-0.0255\subfigwidth]{fig:task2-pos}%
    \begin{minipage}[c]{\dimexpr\columnwidth-\sublabelwidth\relax}
        \centering
        \includegraphics[width=\linewidth]{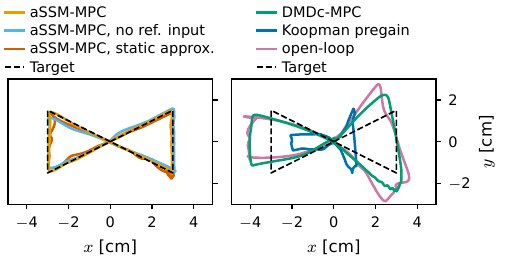}
    \end{minipage}

    \sidesublabel[0.0552\subfigwidth]{fig:task2-angle}%
    \begin{minipage}[c]{\dimexpr\columnwidth-\sublabelwidth\relax}
        \centering
        \includegraphics[width=\linewidth]{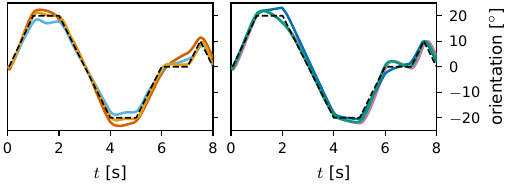}
    \end{minipage}
    \vspace{-0.12cm}
    \caption{Position (a) and orientation (b) of the end effector in task 2. Orientation is plotted as the angle $\arctan(z_5/z_4)$.}
    \label{fig:task2}
    \vspace{-0.15cm}
\end{figure}

The position target in task 1 is a circle of radius 4 [cm] in the $x-y$ plane, reached on a straight path from the unforced steady state, as shown in Figure~\ref{fig:task1-pos}. The simultaneous orientation target can be seen in Figure~\ref{fig:task1-angle}, where the angle $\arctan(z_5 / z_4)$ is plotted for convenience. Task 2 is shown similarly in  Figure~\ref{fig:task2}.

On these tasks, we compare our MPC method to the static-approximation-based aSSM-MPC scheme of \cite{Kaundinya_Alora_Matt_Pabon_Pavone_Haller_2026}, a similar MPC implementation that uses a linear model obtained via DMDc \cite{Proctor_Brunton_Kutz_2016} and the same reference input $u_\textrm{ref}(t)=S^\dagger(\Gamma(t))$ as our method, and the Koopman pregain method from \cite{Haggerty_Banks_Kamenar_Cao_Curtis_Mezic_Hawkes_2023}. The DMDc model is on delay-embedded observables with one delay; increasing this did not improve tracking accuracy in our experiments. The Koopman pregain strategy consists of LQR with the DMDc model and a linear pregain term.

To better understand the differences between our aSSM-MPC and that of \cite{Kaundinya_Alora_Matt_Pabon_Pavone_Haller_2026}, we also test our controller without a reference input, as is done in \cite{Kaundinya_Alora_Matt_Pabon_Pavone_Haller_2026}, and attempt the tasks with the open-loop application of $u(t)=S^\dagger(\Gamma(t))$.

We use a planning horizon of $N=20$ and weight matrices $Q = 100\cdot \diag(I_2, 0.5, 1.5\cdot I_2)$ and $Q_f= 400\cdot \diag(I_2, 0.5, 1.5\cdot I_2)$ wherever applicable. On the other hand, since the input variables play different roles in the different methods, the input and slew rate weights $R$ and $R_{du}$ are tuned individually to optimize performance while keeping oscillations to a minimum. These are shown in Table~\ref{tab:performance}.

\begin{table*}[t]
    \centering
    \setlength{\extrarowheight}{0.5pt}
    \begin{tabular}{|c|c|c|c|c|c|c|}
        \hline
         Method & \multicolumn{2}{|c|}{controller weights}& \multicolumn{2}{|c|}{control effort}& \multicolumn{2}{|c|}{relative $\ell_2$ error} \\
         \hline
         & $R$ & $R_{du}$ & task 1 & task 2 & task 1 & task 2 \\
         \hline
         \hline
         aSSM-MPC & $\diag(0.02, 0.004 \cdot I_6, 2 \cdot I_7)$ & $\diag(10\cdot I_7, 0.1 \cdot I_7)$ & 39.6 & 33.4 & $\mathbf{2.6\%}$ & $\mathbf{2.4\%}$   \\
         \hline
         aSSM-MPC, no reference input & $\diag(0.02, 0.004 \cdot I_6, 2 \cdot I_7)$ & $\diag(10\cdot I_7, 0.1 \cdot I_7)$ & 32.9 & 25.9 & $4.3\%$ & $5.3\%$ \\
         \hline
         aSSM-MPC, static approx. & $0.1 \cdot I_7$ & $0.5\cdot I_7$ & 41 & 35.7 & $6.6\%$ & $6.1\%$ \\
         \hline
         DMDc-MPC & $\diag(0.2, 0.04 \cdot I_6)$ & $10 \cdot I_7$& $41$ & $38.1$ & $14.7\%$  & $16.3\%$  \\
         \hline
         Koopman pregain & $0.12\cdot I_7$ & $-$ & $35.5$ & $21.7$ & $29.9\%$  & $35\%$ \\
         \hline
         open-loop & $-$ & $-$ & 41 & 37.8 & $ 18 \%$  & $23.5\%$ \\
         \hline
    \end{tabular}
    \caption{Parameters, control effort, and tracking performance of the tested controllers.}
    \label{tab:performance}
    \vspace{-0.1cm}
\end{table*}

\begin{figure}
    \centering
    \sidesublabel[-0.0618\subfigwidth]{fig:error-task1}%
    \begin{minipage}[c]{\subfigwidth}
        \centering
        \includegraphics[width=\linewidth, trim={0 76.5 0 0}, clip]{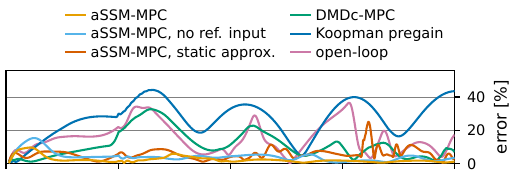}
    \end{minipage}

    \sidesublabel[0.0500\subfigwidth]{fig:error-task2}%
    \begin{minipage}[c]{\subfigwidth}
        \centering
        \includegraphics[width=\linewidth, trim={0, 0, 0, 82}, clip]{instant_error_both_tasks_rightaxis.pdf}
    \end{minipage}
    \vspace{-0.15cm}
    \caption{Instantaneous relative error in tasks 1 (a) and 2 (b).}
    \label{fig:both-tasks-error}
    \vspace{-0.25cm}
\end{figure}

The performance of each control strategy is reported in Table~\ref{tab:performance}, which shows the norm of each error, normalized by that of the target, in terms of the finite-horizon signal $\ell_2$-norm $\norm{x}_{\ell_2, T_\Gamma} = \big(\sum_{t_k=0}^{T_\Gamma} \norm{x(t_k)}_2^2\big)^{1/2}$. 
The trajectories are plotted in Figures~\ref{fig:task1} and \ref{fig:task2}, while instantaneous relative error comparisons are given in Figure~\ref{fig:both-tasks-error}.

Observe that our aSSM-MPC method achieves high tracking accuracy and outperforms the DMDc-based alternatives by a large margin. Although to a lesser extent, our modifications also lead to meaningful performance gains over MPC with static aSSM-approximations, especially in orientation tracking and at larger distances from the unforced equilibrium. The poor performance of the open-loop application of the reference input and the increased errors caused by removing the reference input from our aSSM-MPC strategy underscore the advantage of using our proposed modifications in combination.

For context, we also report the control effort of each run in Table~\ref{tab:performance}, computed as the signal $\ell_2$-norm of the applied input sequence. Only two methods, the ablation without a reference input and Koopman pregain, have lower control effort than ours. The former deliberately has the same weights as our full aSSM-MPC method and the lower control effort follows from applying the regularization to the slow input $u^s$ itself instead of its deviation from $u_\textrm{ref}$. For the latter, increasing the control effort by tuning the weights, while maintaining numerical stability, was not possible in our experiments.

Across the two tasks, the total compute time per MPC iteration (excluding the first time step) of our method had a mean of 4.85 [ms], maximum of 9.45 [ms], and 95th percentile at 7.38 [ms], measured on a MacBook Pro with a 15-core M5 Pro chip and 24 GB RAM. That is, for a 100 [Hz] discretization, it runs in real time.

Lastly, we note that the slowness metric of task 1 is $r_s=2.7$ and that of task 2 is $r_s=2$. This suggests that aSSM-reduced models are applicable in practice even when $r_s<<1$ no longer holds.

\section{Conclusion}

We have extended the aSSM-MPC scheme of \cite{Kaundinya_Alora_Matt_Pabon_Pavone_Haller_2026} to use the full, time-varying aSSM geometry for predictions and to optimize over both the slow and the fast input components, enabling more accurate predictions and greater control authority. In addition, we have introduced a set of orientation observables for a soft arm, and developed aSSMPy, a Python implementation of data-driven aSSM-reduction.

Combining these, we learned a seven-dimensional aSSM-reduced model of the soft arm SoPrA and deployed our aSSM-MPC scheme to track dynamic position and orientation targets in a high-fidelity finite-element simulator. In both test tasks, our approach achieved high tracking accuracy, outperforming the existing aSSM-MPC of \cite{Kaundinya_Alora_Matt_Pabon_Pavone_Haller_2026} and, by an especially large margin, linear baselines.

Since our methods are entirely data-driven and require a training dataset of modest size, 
they are well suited for direct deployment on hardware. Thus, validating the approach on the physical soft arm is an immediate next step and has already been successful for a similar method on a tendon-driven robot in ongoing work \cite{Kaundinya_et_al_2026}. Another  important direction remaining is the derivation of rigorous stability guarantees for aSSM-MPC.

\vspace{-0.075cm}
\section*{Acknowledgments}
The authors used suggestions by Claude Opus 5 to improve specific segments of the text.
\vspace{-0.075cm}

\bibliography{IEEEabrv,references}
\bibliographystyle{IEEEtran}
\end{document}